\documentclass[letterpaper, 10 pt, journal, twoside]{ieeetran} 
\usepackage{ifpdf}
\usepackage{algorithmic}
\usepackage{fixltx2e}
\usepackage[T1]{fontenc}
\usepackage[utf8]{inputenc}
\usepackage{cite}
\usepackage{amsmath,amsfonts}
\usepackage{amssymb}
\usepackage{array}
\usepackage{textcomp}
\usepackage{url}
\usepackage{verbatim}
\usepackage{graphicx}
\usepackage{graphics} 
\usepackage{mathtools}
\usepackage{booktabs}
\usepackage{adjustbox}
\usepackage{stfloats}
\usepackage[breaklinks=true,colorlinks,citecolor=blue]{hyperref}
\usepackage{rotating}
\usepackage{subfloat}
\usepackage{makecell}
\usepackage{multirow}
\usepackage{balance}
\usepackage{academicons}
\usepackage{xcolor}
\usepackage{lipsum}
\usepackage{colortbl}
\usepackage{rotating}
\usepackage[linesnumbered,ruled,vlined,commentsnumbered]{algorithm2e}
\usepackage[caption=false, font=footnotesize]{subfig}
\usepackage{footnote}
\usepackage{tabularray}
\usepackage{tablefootnote}
\usepackage[normalem]{ulem}
\UseTblrLibrary{siunitx}
\usepackage{amssymb}
\usepackage{pifont}
\newcommand{\xmark}{\ding{55}}  
\newcommand{\nmark}{\ding{52}}  

\begin{document}
%
\title{\textit{Have We Scene It All?} Scene Graph-Aware\\ Deep Point Cloud Compression}

\makeatletter
\newcommand*\titleheader[1]{\gdef\@titleheader{#1}}
\AtBeginDocument{%
  \let\st@red@title\@title
  \def\@title{%
    \vskip-2.0em
    \bgroup\normalfont\small\centering\@titleheader\par\egroup
    \vskip1.5em\st@red@title}
}
\makeatother

\titleheader{Please cite as: N. Stathoulopoulos, C. Kanellakis and G. Nikolakopoulos, "\textit{Have We Scene It All?} Scene Graph-Aware Deep Point Cloud Compression," in IEEE Robotics and Automation Letters, vol. 10, no. 12, pp. 12477-12484, 2025, doi: 10.1109/LRA.2025.3623045.}

%
%
%

\author{Nikolaos Stathoulopoulos, Christoforos Kanellakis and George Nikolakopoulos%
\thanks{This paper was supported in part and received funding from the European Union’s Horizon Europe Research and Innovation Programme under the Grant Agreement No.101138451.} 
\thanks{The authors are with the Robotics and AI Group, Department of Computer, Electrical and Space Engineering, Lule\r{a} University of Technology, 971 87 Lule\r{a}, Sweden.{Corresponding Author's e-mail: \tt\footnotesize niksta@ltu.se}}%
\thanks{The code implementation and pre-trained model weights of this letter are available at: \texttt{\href{https://github.com/LTU-RAI/sga-dpcc}{https://github.com/LTU-RAI/sga-dpcc.git}}.}
}
\maketitle

\begin{abstract}
Efficient transmission of 3D point cloud data is critical for advanced perception in centralized and decentralized multi-agent robotic systems, especially nowadays with the growing reliance on edge and cloud-based processing.    
However, the large and complex nature of point clouds creates challenges under bandwidth constraints and intermittent connectivity, often degrading system performance. 
We propose a deep compression framework based on semantic scene graphs. The method decomposes point clouds into semantically coherent patches and encodes them into compact latent representations with semantic-aware encoders conditioned by Feature-wise Linear Modulation (FiLM). A folding-based decoder, guided by latent features and graph node attributes, enables structurally accurate reconstruction. 
Experiments on the SemanticKITTI and nuScenes datasets show that the framework achieves state-of-the-art compression rates, reducing data size by up to 98\% while preserving both structural and semantic fidelity. In addition, it supports downstream applications such as multi-robot pose graph optimization and map merging, achieving trajectory accuracy and map alignment comparable to those obtained with raw LiDAR scans.
\end{abstract}


%
\IEEEpeerreviewmaketitle

\section{Introduction}
\IEEEPARstart{I}{n} recent years, robotic perception has shifted from image‐based representations to rich, three‐dimensional point clouds, driven by the need for greater robustness in complex, unstructured and ill-illuminated environments. At the same time, multi‐robot systems are proliferating, and both centralized coordination and peer‐to‐peer collaboration increasingly rely on transmitting 3D data to edge or cloud servers for joint processing~\cite{seisa2022edge}.
The advent of high‐bandwidth 5G networks and edge‐computing platforms promises lower latency and higher throughput~\cite{damigos2024-5G}, yet the sheer volume and irregular structure of point clouds still impose prohibitive demands on storage, transmission, and real‐time responsiveness.

Point clouds are fundamentally large since each sweep can contain thousands of spatial samples and they lack the regular grid structure that makes images compliant to classical compression schemes. Under realistic bandwidth constraints and with intermittent connections, 
naive streaming of full‐resolution clouds leads to dropped frames, degraded map accuracy~\cite{damigos2023com_aware}, and ultimately failures in perception‐driven tasks such as obstacle avoidance or cooperative mapping~\cite{stathoulopoulos2024frame}. While semantic scene graphs have proven effective as compact, task‐oriented abstractions for navigation and planning
existing compression pipelines rarely exploit their structural information to guide lossy encoding of raw geometry.
\begin{figure}[!t]
    \centering
    \includegraphics[width=1.0\linewidth]{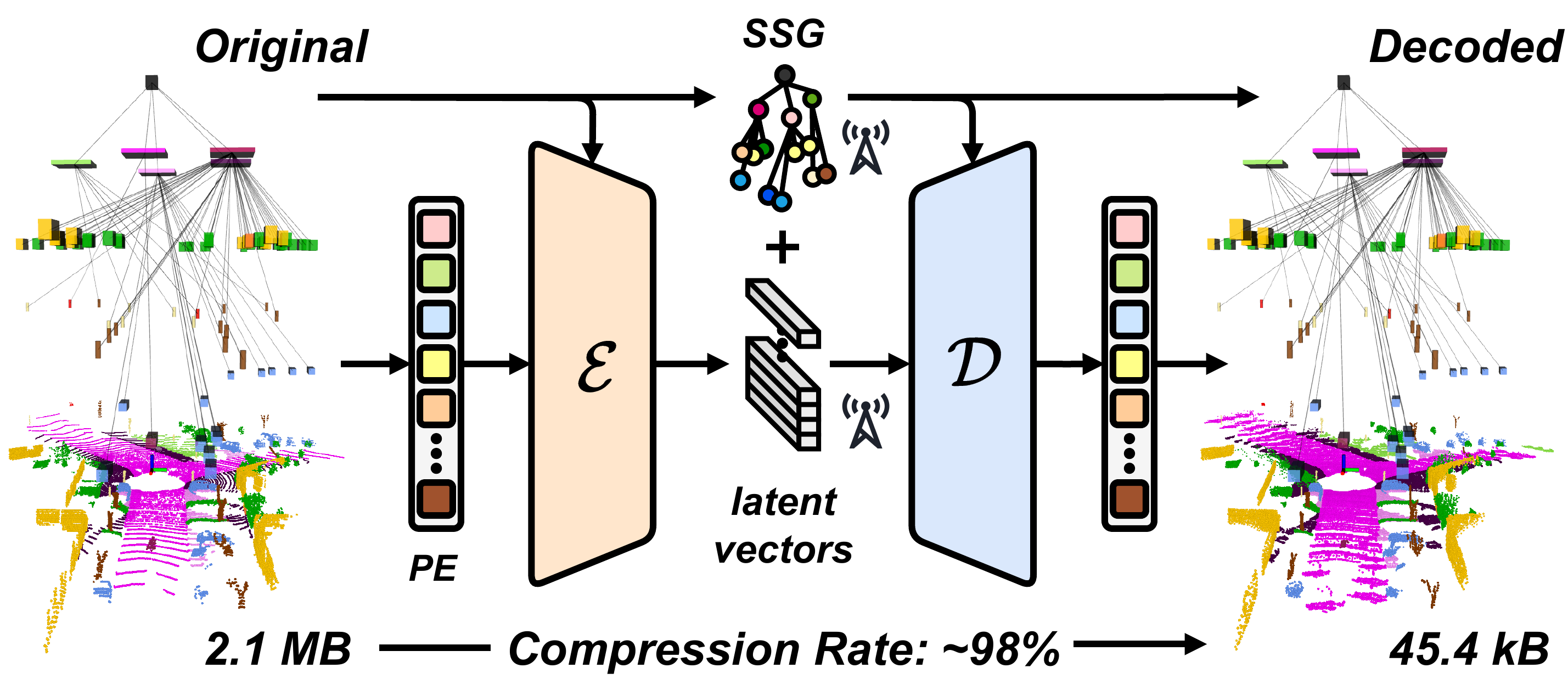}
    \setlength{\abovecaptionskip}{-15pt}
    \caption{\textbf{Overview.} A raw point cloud is first converted into a semantic scene graph (SSG), capturing object- and layer-level structure. The patch extractor (PE) then subdivides the scene into layer-specific patches, which are encoded by a transformer-based autoencoder into compact, per-patch latent vectors. These are later decoded to reconstruct the full point cloud. The proposed framework achieves extreme compression rates of up to 98\%, with the encoded representation consisting solely of the scene graph and the set of latent vectors.}
    \vspace{-0.65cm}
    \label{fig:concept}
\end{figure}

While previous works demonstrate the advantages of embedding relational and semantic information in 3D representations, they focus on mapping, localization~\cite{dube2020segmap}, or scene synthesis~\cite{armeni20193d,gao2023scenehgn} rather than data compression. To our knowledge, no existing method has systematically fused graph- or structure-aware abstractions with point cloud compression. In this letter, we close this gap by introducing a scene graph–conditioned autoencoder for point cloud compression. Our framework learns compact latent vectors that jointly preserve geometric fidelity and relational structure, yielding a unified representation that not only enables highly efficient compression but could also further support downstream tasks such as semantic scene understanding, place recognition and beyond.

In more detail, we propose a novel deep compression framework that combines semantic scene graphs with patch‐based autoencoding. First, a semantic scene graph generator partitions each LiDAR scan into node patches, each corresponding to a coherent object, agent, infrastructure element, or terrain segment. Each patch is then fed into a semantic‐aware encoder where point embeddings and positional encodings are modulated via Feature-wise Linear Modulation (FiLM) layers conditioned on learned class embeddings, and a transformer backbone distills the points into a compact latent vector. On the decoder side, a folding‐based upsampler, conditioned on the node’s bounding box attributes, reconstructs a high‐fidelity point set from each encoded latent vector. By transmitting only these latents and the minimal metadata of the graph structure, our method drastically reduces the communication load.

In sum, the main contributions of this letter are:
(a) A graph‐conditioned autoencoder for deep point cloud compression. We introduce a novel framework that decomposes raw LiDAR scans into semantically coherent patches using a semantic scene graph, and encodes each patch via a FiLM‐conditioned transformer to jointly compress geometric and semantic cues.
(b) Integration of relational structure into compression. Our method leverages semantic scene graphs throughout the pipeline, using semantic conditioning in the encoder and attribute-guided folding in the decoder to preserve relational consistency in the compressed representation.
(c) Extensive experiments on SemanticKITTI and nuScenes demonstrate state-of-the-art performance, achieving up to 98\% data reduction while preserving both geometric and semantic fidelity. Our method outperforms leading codecs such as Draco~\cite{draco} and the MPEG approach by Mekuria et al.~\cite{mekuria2017mpeg}, and maintains strong performance in downstream robotic tasks such as multi-agent pose graph optimization and map merging, even under strict bandwidth constraints.
\vspace{-0.25cm}

\section{Related Work}

In this section, we split the related work in two areas. First, we discuss methods for point cloud compression, from classical spatial partitioning to recent learning-based approaches. Second, we highlight works that incorporate relational and structural information into 3D point cloud representations, motivating our scene graph-aware compression framework.

\textbf{Point Cloud Compression:}  
Classical methods based on spatial partitioning, such as Octrees~\cite{octattention}, provide compact representations by recursively subdividing space into occupied and free regions. While effective at reducing memory usage, they often fail to preserve fine-grained geometry at high compression ratios. Beyond this, several approaches exploit structural redundancies.  
Sun et al.~\cite{Sun2019clustering} proposed a clustering-based method that segments range images into ground and object regions, followed by region-specific predictive encoding. 
Feng et al.~\cite{Feng2020} introduced a spatio-temporal compression framework combining keyframe selection and iterative plane fitting to model static structures across sequences.

More recently, learning-based approaches have further enhanced compression performance.  
Wiesmann et al.~\cite{wiesmann2021deep} presented a deep convolutional autoencoder that directly operates on dense point cloud maps, using Kernel Point Convolutions ({KPConv}~\cite{thomas2019KPConv}) for feature extraction and a deconvolution operator for flexible upsampling.
Voxel-based and sparse tensor methods have also gained traction, with examples like {VoxelContext-Net}~\cite{Que2021VoxelContext}, that augment octree encoding with voxel context modeling through deep entropy estimation, or {SparsePCGC}~\cite{Wang2023tensor} which introduced a multiscale sparse tensor framework focused on the most probable occupied voxels.

Regarding range image-based methods,
{FLiCR}~\cite{Heo2022FLiCR} compresses 3D LiDAR data by projecting it into 2D range images and applying lossy quantization and downsampling, enabling lightweight real-time operation on resource-constrained platforms.  
{RIDDLE}~\cite{Theis2022riddle} improves compression by modeling local differences between range image pixels through deep delta encoding, significantly reducing redundancy while preserving fine structural details.  
Moreover, You et al. proposed RENO~\cite{you2025reno}, a real-time neural compression framework that leverages sparse tensor representations for efficient geometry encoding, achieving high compression efficiency while maintaining fast inference suitable for time-critical applications.
Finally, semantic-aware compression methods~\cite{Zhao2022semantic} have also emerged, with scene priors integrated into the compression pipeline, jointly optimizing for geometric fidelity and downstream perception tasks.

While most compression approaches focus on geometric preservation and memory reduction, some recent works explore compression jointly with downstream applications.  
{RecNet}~\cite{stathoulopoulos2024recnet} proposes an invertible point cloud encoding via range image embeddings, enabling efficient multi-robot map sharing, reconstruction and place recognition. This trend towards combining compression with task-driven objectives highlights the potential of designing representations that are not only compact but also structurally and semantically informative.

\textbf{Graph-- and Structure--Aware Representations:}  
Beyond compression, several works focus on enriching point cloud representations with structural, relational, and semantic information.  
{SegMap}~\cite{dube2020segmap} introduced a segment-based mapping framework, partitioning 3D LiDAR point clouds into local segments and encoding them with compact learned descriptors. While it demonstrates the benefits of structure-aware feature learning for tasks such as global localization and 3D reconstruction, it primarily focuses on segment-level representations and does not explicitly model the relational structure between objects. 
Other works pursued explicit scene graph construction from 3D data.  
Armeni et al.~\cite{armeni20193d} proposed 3D Scene Graphs that organize environments into hierarchical semantic and geometric structures, while Gao et al.~\cite{gao2023scenehgn} developed hierarchical scene graphs for indoor scene synthesis, highlighting the role of relational context in modeling complex spaces.

At the network architecture level, point cloud processing has evolved toward dynamic, graph-based methods. 
{DGCNN}~\cite{wang2019dynamic} introduced dynamic edge convolution operations that adaptively capture local geometric structures during learning, while {RandLA-Net}~\cite{hu2020randla} proposed lightweight local feature aggregation schemes for large-scale outdoor scenes. These advances emphasize the advantages of relational inductive biases for 3D understanding, while object-centric mapping approaches further demonstrate the value of structure-aware modeling.  

Existing works focus on scene-aware representations for perception and mapping, but not compression. We leverage scene graph organization during compression to structure the encoding process according to semantic and geometric layers, enabling more efficient and consistent latent representations.
\vspace{-0.15cm}

\section{The Proposed Framework}

Our goal is to compress a point cloud into compact latent representations, significantly reducing data size, while preserving structural and semantic integrity for efficient transmission between robotic agents. 
Given a point cloud $\mathcal{P} \subset \mathbb{R}^3$, we first apply a semantic segmentation mapping $S: \mathbb{R}^3 \rightarrow \mathbb{R}^4$, yielding a labeled point cloud $\bar{\mathcal{P}} = S(\mathcal{P}) \subset \mathbb{R}^4$. We then construct a semantic scene graph, decomposing the cloud into semantically coherent clusters (nodes). Each node is then independently encoded by a semantic-aware encoder conditioned via FiLM~\cite{perez2018film}, generating compact latent vectors. These latent vectors, combined with the scene graph, guide a folding-based decoder to accurately reconstruct the original point cloud.

\subsection{Semantic Scene Graph (SSG) Generation} \label{subsec:ssg_gen}

Given the labeled point cloud $\bar{\mathcal{P}}$, we construct a layered 3D semantic scene graph $\mathcal{G} = (\mathcal{V}, \mathcal{E})$, capturing both semantic and spatial relationships. Each node $v_i \in \mathcal{V}$ corresponds to a semantic cluster defined by its class $l_i$, spatial center $\boldsymbol{c}_i \in \mathbb{R}^3$, extent $\boldsymbol{e}_i \in \mathbb{R}^3$, and orientation $\boldsymbol{R}_i \in SO(3)$.
We partition the graph into hierarchical layers: (1) \textit{terrain}, (2) \textit{infrastructure}, (3) \textit{objects}, and (4) \textit{agents}. Nodes are created using their class labels and their geometric structure through well-established methods, such as plane or cylinder fitting, region growing, or using instance labels if available.
Edges $(v_i, v_j) \in \mathcal{E}$ represent proximity-based or hierarchical connections across layers. Specifically, each frame \textit{(layer 0)} connects to terrain nodes \textit{(layer 1)}, encompassing road, sidewalk, parking, grass and so on. Subsequent layers \textit{(infrastructure}, \textit{objects}, \textit{agents)} connect to terrain nodes based on spatial proximity. For example, moving vehicles typically connect to road nodes, parked vehicles to parking nodes, and vegetation or signs to grass or sidewalk nodes. Nodes without clear terrain associations (e.g., buildings or bushes) connect to a generic ``other'' \textit{terrain} node. While these connections do not directly influence the compression pipeline's performance, they provide a structured context beneficial for downstream tasks such as map filtering.
\vspace{-0.7cm}

\subsection{Autoencoder Architecture}

Given a semantic point cloud $\bar{\mathcal{P}}_{t}$ and its corresponding semantic scene graph $\mathcal{G}_t$, our goal is to compress each semantic node independently using a dedicated, layer-specific autoencoder. Each autoencoder comprises a semantic-aware encoder and a scene graph-conditioned decoder, trained end-to-end with a multi-component loss function. This ensures compact, semantically consistent latent representations.

\subsubsection{Patch Extraction}

To maintain compactness for transmission, 3D points are not stored directly within the graph. Instead, given a node $v_i \in \mathcal{V}$ of the semantic scene graph, we retrieve the corresponding point cloud patch $\pi_i$ from the labeled point cloud $\bar{\mathcal{P}}$. Specifically, for \textit{non-terrain} nodes, extraction is straightforward, and each node corresponds to a single patch ($v_i \equiv \pi_i$), defined by the node's oriented bounding box with center $\boldsymbol{c}_i$, extent $\boldsymbol{e}_i$, and orientation $\boldsymbol{R}_i$:
\begin{equation}
    \pi_i = \{\boldsymbol{x}_j \in \bar{\mathcal{P}} \mid l_j = l_i,\: \boldsymbol{R}_i^\top(\boldsymbol{x}_j - \boldsymbol{c}_i) \in [-\tfrac{\boldsymbol{e}_i}{2}, \tfrac{\boldsymbol{e}_i}{2}]\}.
\end{equation}
Conversely, for \textit{terrain} nodes, which typically represent larger spatial extents (e.g., road segments), we subdivide the area into multiple smaller patches $\{\pi_{i,k}\}$, ensuring uniform point distribution and sufficient geometric detail. The set of these smaller patches collectively reconstruct the original node, $v_i = \{\pi_{i,1}, \pi_{i,2}, \dots, \pi_{i,K}\}$.
For efficient computation, all extracted patches are uniformly subsampled or padded with masking to a fixed maximum number of points $N_{\text{layer}}$ (unique to each layer), ensuring consistent input shapes for downstream processing.

\subsubsection{Encoder}
\begin{figure}[!t]
    \centering
    \includegraphics[width=0.75\linewidth]{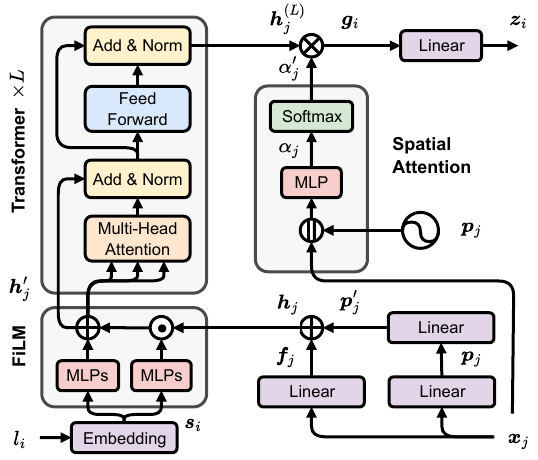}
    \setlength{\abovecaptionskip}{-2pt}
    \caption{\textbf{Semantic-aware Encoder.} Overview of the proposed semantic-aware encoder, where each patch is processed alongside its semantic class. The right section illustrates the positional encoding module, which maps 3D coordinates to a high-dimensional space and projects them into the feature space. A FiLM module conditions point features using the semantic embedding of the patch, enabling the network to adapt representations based on semantic context. These features are then refined through a series of transformer blocks and pooled via spatial attention to produce a compact latent descriptor.}    
    \vspace{-0.4cm}
    \label{fig:encoder}
\end{figure}

Given a patch $\pi_i \in \mathbb{R}^{N \times 3}$ and the corresponding semantic label $l_i$, the encoder, as seen in Fig.~\ref{fig:encoder}, maps it to a latent vector $\boldsymbol{z}_i = \mathcal{E}(\pi_i, l_i)$ through the following stages.

\noindent\textbf{Point-wise Encoding:}  
Each point $\boldsymbol{x}_j \in \mathbb{R}^3$ is first mapped to a higher-dimensional feature space $\mathbb{R}^{D_f}$ via a point-wise Linear layer $\Lambda_\theta^f$, yielding: $\boldsymbol{f}_j = \Lambda_\theta^f(\boldsymbol{x}_j), \: \text{where} \:\: \boldsymbol{f}_j \in \mathbb{R}^{D_f} \:\: \text{and} \:\: j = 1, \dots, N.$
Here $\theta$ are the trainable parameters of the embedding.
In parallel, each point is passed through a separate Linear layer $\Lambda_\theta^p$ to obtain a positional embedding $\boldsymbol{p}_j \in \mathbb{R}^{D_p}$, which is then projected into the same feature space with a second projection layer $\Lambda_\theta^{p'}: \mathbb{R}^{D_p} \rightarrow \mathbb{R}^{D_f}$: $\boldsymbol{p}_j' = \Lambda_\theta^{p'}\big(\Lambda_\theta^p(\boldsymbol{x}_j)\big).$
The final encoded representation per point is the sum of feature and positional embeddings: $\boldsymbol{h}_j = \boldsymbol{f}_j + \boldsymbol{p}_j', \:\: \text{where} \:\: \boldsymbol{h}_j \in \mathbb{R}^{D_f}.$

\vspace{0.33em}
\noindent\textbf{Semantic Conditioning via {FiLM}:}  
To inject semantic context, we embed the node-level semantic label $l_i$ using an Embedding layer $E_\theta^s$, producing a semantic descriptor: $\boldsymbol{s}_i = E_\theta^s(l_i), \:\: \text{where} \:\: \boldsymbol{s}_i \in \mathbb{R}^{D_s}.$
We condition the per-point features using Feature-wise Linear Modulation ({FiLM})~\cite{perez2018film}, which applies a learned affine transformation to each feature dimension: $\boldsymbol{h}_j' = \gamma(\boldsymbol{s}_i) \odot \boldsymbol{h}_j + \beta(\boldsymbol{s}_i),$
where $\gamma(\cdot)$ and $\beta(\cdot)$ are \texttt{MLPs} that output scaling and shifting vectors respectively, and $\odot$ denotes element-wise multiplication.
This conditioning enables the encoder to adapt its representation based on semantic class, improving generalization across heterogeneous patches.

\vspace{0.33em}
\noindent\textbf{Transformer-based Encoding:}  
The semantically conditioned features $\boldsymbol{H}^{(0)} = [\boldsymbol{h}'_1, \dots, \boldsymbol{h}'_N]$ are processed by a sequence of $L$ Transformer blocks, each comprising Multi-Head Self-Attention (\texttt{MH-Attn}), Feed-Forward Networks ($\Phi$), and residual connections with Layer Normalization (\texttt{Norm}). Formally, at each layer $\ell = 1, \dots, L$, the update is computed as:
\begin{align}
    \boldsymbol{H}' \:\:\: = \ &\texttt{Norm}\big(\boldsymbol{H}^{(\ell-1)} + \texttt{Drop}\big(\texttt{MH-Attn}\big(\boldsymbol{H}^{(\ell-1)}\big)\big)\big), \\    
    \boldsymbol{H}^{(\ell)} = \ &\texttt{Norm}\big(\boldsymbol{H}' + \texttt{Drop}\big(\Phi_\theta(\boldsymbol{H}')\big)\big),
\end{align}
The Multi-Head Self-Attention module~\cite{attentionTrans} computes attention across all point features, enabling rich contextual encoding of intra-patch geometric relationships. Residual connections and dropout are applied at both attention and Feed-Forward stages to improve stability and generalization.

\vspace{0.33em}
\noindent\textbf{Spatial-Attention Pooling:}  
To aggregate point-wise features into a global patch descriptor, we employ Spatial-Attention pooling. For each point, we compute an attention score via a lightweight \texttt{MLP}, taking as input the concatenation of coordinates and positional encodings: $\alpha_j = \texttt{MLP}_\theta^\alpha(\boldsymbol{x}_j \mathbin\Vert \boldsymbol{p}_j) \in \mathbb{R}.$
The attention weights are then normalized using softmax:
\begin{equation}
    \alpha_j' = \frac{\exp(\alpha_j)}{\sum_{k=1}^N \exp(\alpha_k)}, \:\: \text{where} \:\: j = 1, \dots, N.
\end{equation}
The global feature vector $\boldsymbol{g}_i \in \mathbb{R}^{D_f}$ is computed as a weighted sum of the final transformer outputs: $\boldsymbol{g}_i = \sum_{j=1}^N \alpha_j' \boldsymbol{h}_j^{(L)}.$

\vspace{0.33em}
\noindent\textbf{Latent Representation:}  
The aggregated global feature is projected to a compact latent vector through a final Linear layer: $\boldsymbol{z}_i = \Lambda_\theta^z(\boldsymbol{g}_i) \in \mathbb{R}^{D_z}$. This latent representation captures both the geometric and semantic context of the input patch, enabling precise reconstruction by the following decoder.

\subsubsection{Decoder}
\begin{figure}[!t]
    \centering
    \includegraphics[width=0.75\linewidth]{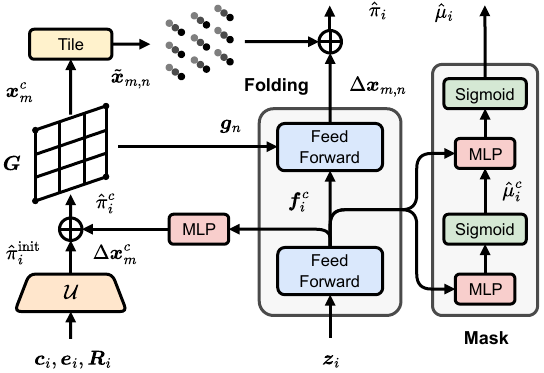}
    \setlength{\abovecaptionskip}{-3pt}
    \caption{\textbf{Scene graph-conditioned Decoder.} Overview of the proposed decoder conditioned on the semantic scene graph attributes. Given the latent vector of a patch, the decoder generates a set of coarse points using the bounding box and learns per-point offsets. These coarse features are then upsampled via a folding operation guided by a fixed 2D grid, producing a dense reconstruction. A final confidence mask is predicted to prune low-quality outputs.}
    \vspace{-0.45cm}
    \label{fig:decoder}
\end{figure}

The decoder reconstructs a point cloud patch from the latent vector $\boldsymbol{z}_i \in \mathbb{R}^{D_z}$, conditioned on the node's geometric parameters: center $\boldsymbol{c}_i$, extent $\boldsymbol{e}_i$, and orientation $\boldsymbol{R}_i$. As input patches are padded to a fixed size $N$ for batching, the decoder also predicts a confidence mask to indicate valid points, accounting for sparsity due to LiDAR range.
We define the decoder function, as seen in Fig.~\ref{fig:decoder}, as: $\hat{\pi}_i, \hat{\mu}_i = \mathcal{D}(\boldsymbol{z}_i, \boldsymbol{c}_i, \boldsymbol{e}_i, \boldsymbol{R}_i)$,
where $\hat{\pi}_i \in \mathbb{R}^{N \times 3}$ is the reconstructed patch, and $\hat{\mu}_i \in [0,1]^N$ is the per-point confidence mask.

\vspace{0.33em}
\noindent\textbf{Coarse Point Generation:}
We begin by generating $M$ coarse patch points $\hat{\pi}_i^{\text{init}} = \{\boldsymbol{x}_m^\text{init}\}$ within the node's bounding box, where each point is sampled using a uniform distribution $\mathcal{U}\big(\boldsymbol{c}_i, \boldsymbol{e}_i, \boldsymbol{R}_i \big).$
Next, the latent vector $\boldsymbol{z}_i$ is mapped to coarse point features via a feed-forward network comprising a two-layer \texttt{MLP} with ReLU activations: $\boldsymbol{f}_i^{{c}} = \Phi_{\theta}^{{c}}(\boldsymbol{z}_i), \, \boldsymbol{f}_i^{{c}} \in \mathbb{R}^{M \times D_f^c},$
where $D_f^c$ denotes the dimensionality of the associated coarse features, and $\boldsymbol{f}_i^{c} = \{\boldsymbol{f}_m^c\}_{m=1}^M$ with each $\boldsymbol{f}_m^c \in \mathbb{R}^{D_f^c}$ representing the feature of the $m$-th coarse point. These features are then processed by a separate \texttt{MLP}: $\Delta \boldsymbol x_m^{c} = \texttt{MLP}^{\text{offset}}_\theta(\boldsymbol{f}_{m}^{c}), \, m = 1,\dots,M, \, \Delta \boldsymbol{x}_{m}^c \in \mathbb{R}^3,$
allowing the model to learn the rough shape of each patch. This uniform initialization provides a structured set of base coordinates, while the latent-conditioned offsets adapt these points to the true geometry, ensuring stable training and faithful reconstruction.
The coarse patch points $\hat{\pi}_i^c = \{\boldsymbol{x}_m^c\} \in \mathbb{R}^{M \times 3}$ are formed as: $\hat{\pi}_i^c = \big\{ \boldsymbol{x}_m^\text{init} + \Delta \boldsymbol{x}_m^c \mid \boldsymbol{x}_m^\text{init} \in \hat{\pi}_i^\text{init}; \; m = 1,\dots,M \big\}.$
Additionally, we predict a coarse confidence mask $\hat{\mu}_i^{c}$ indicating the validity of each coarse point via a small \texttt{MLP} followed by a sigmoid activation function $\sigma$: $\hat{\mu}_i^{c} = \sigma\left(\texttt{MLP}_\theta^{\text{mask}}(\boldsymbol{f}_i^{c})\right), \, \hat{\mu}_i^{c} \in [0,1]^M.$
This coarse patch defines the initial structure upon which the fine reconstruction is built.

\vspace{0.33em}
\noindent\textbf{Fine Reconstruction via Folding:}  
To produce a higher resolution output of size $N$, we employ a folding-based upsampling strategy~\cite{foldingnet} that refines each of the $M$ coarse points using a structured 2D grid of $G$ positions, such that $N = M \cdot G$. Each coarse point is thus expanded into a local neighborhood of fine points through learned geometric offsets.
Given the set of coarse patch points $\hat{\pi}_i^c$ and their associated features $\boldsymbol{f}_i^c = \{\boldsymbol{f}_m^c\}$, we define a folding grid $\boldsymbol{G}(\hat{\pi}_i^c) = \{\boldsymbol{g}_n\}_{n=1}^{g \times g} \in \mathbb{R}^2$, where $G = g^2$ and each $\boldsymbol{g}_n$ is a 2D coordinate.
For each pair of coarse features and grid locations, a shared feed-forward network predicts a 3D offset: $\Delta \boldsymbol{x}_{m,n} = \Phi_{\theta}^{\text{fold}}(\boldsymbol{f}_m^c, \boldsymbol{g}_n), \, \Delta \boldsymbol{x}_{m,n} \in \mathbb{R}^3,$
where $m = 1, \dots, M$ and $n = 1, \dots, G$.
Then, we tile the grid $\boldsymbol{G}$ around each coarse point by defining base coordinates: $\tilde{\boldsymbol{x}}_{m,n} = \boldsymbol{x}_m^c, \, \forall \,\, n = 1,\dots,G,$
and obtain the final fine-level reconstruction by adding the learned offsets: $\hat{\pi}_i = \big\{\tilde{\boldsymbol{x}}_{m,n} + \Delta \boldsymbol{x}_{m,n} \mid m = 1,\dots,M;\; n = 1,\dots,G \big\}$.
Finally, we predict the per-point confidence mask $\hat{\mu}_i$ using another \texttt{MLP} network, conditioned on the coarse features and their associated confidence scores: $\hat{\mu}_i = \sigma(\texttt{MLP}_{\phi}^{\text{mask}}(\boldsymbol{f}_i^c, \hat{\mu}_i^c)), \, \hat{\mu}_i \in [0,1]^N$.
This upsampling process offers a resolution-aware reconstruction of fine-grained geometries, while maintaining semantic coherence through the structured latent representation and bounding box context.
\vspace{-0.4cm}

\subsection{Loss Functions}
We train the autoencoders to reconstruct a point cloud as faithful as possible to the input.
Given the ground-truth patch $\pi_i$ with associated valid point mask $\mu_i$, and the predicted reconstructions $\hat{\pi}_i$ (fine) and $\hat{\pi}_i^c$ (coarse) along with the predicted confidence masks $\hat{\mu}_i$ and $\hat{\mu}_i^c$, the loss is defined as:
\begin{align}
    \mathcal{L} = D_{\text{CD}}(\pi_i, \hat{\pi}_i)\; + &\; \lambda_1 D_{\text{CD}}(\pi_i, \hat{\pi}_i^c) + \lambda_2 \mathcal{L}_d(\pi_i, \hat{\pi}_i) \nonumber \\
    & + \lambda_3 \mathcal{L}_m(\mu_i, \hat{\mu}_i) + \lambda_4 \mathcal{L}_m(\bar{\mu}_i, \hat{\mu}_i^c),
\end{align}
where $\bar{\mu}_i$ denotes a downsampled version of $\mu_i$ aligned to the number of coarse points.
The weights $\lambda_1$ to $\lambda_4$ are scheduled during training, progressively decaying to reduce the influence of the auxiliary regularization terms as optimization focuses more on detailed reconstruction.
The first term enforces accurate fine-level reconstruction, while the remaining terms regularize coarse reconstruction accuracy, spatial uniformity, and confidence mask prediction respectively. Each component is detailed below.

\vspace{0.33em}
\noindent\textbf{Chamfer Distance ($D_\text{CD}$):}
We supervise both the fine-level and coarse-level reconstructions using the Chamfer Distance, which measures the symmetric average of the squared distances between each point and its nearest-neighbor in the other point cloud:

\begin{align} \label{eq:chamfer}
    D_\text{CD}\big(\mathcal{P}_\text{trg}, \mathcal{P}_\text{src}\big) &= \frac{D_e\big(\mathcal{P}_\text{trg}, \mathcal{P}_\text{src}\big)}{2} + \frac{D_e\big(\mathcal{P}_\text{src}, \mathcal{P}_\text{trg}\big)}{2}, \\
    D_e\big(\mathcal{P}_i, \mathcal{P}_j\big) &= \frac{1}{|\mathcal{P}_i|} \sum_{\boldsymbol{x}_i \in \mathcal{P}_i} \min_{\boldsymbol{x}_j \in \mathcal{P}_j} {\underbrace{\|\boldsymbol{x}_i - \boldsymbol{x}_j\|_2^2}_{d(\boldsymbol{x}_i,\boldsymbol{x}_j)}}.  \label{eq:distance}
\end{align}
Here, $\mathcal{P}_\text{trg}$ and $\mathcal{P}_\text{src}$ are the sets of target and source 3D points, and $D_e(\mathcal{P}_i, \mathcal{P}_j)$ denotes the mean squared nearest-neighbor distance from $\mathcal{P}_i$ to $\mathcal{P}_j$.

\vspace{0.33em}
\noindent\textbf{Density Regularization ($\mathcal{L}_d$):}
To encourage uniform spatial coverage and improve structural similarity, we introduce a voxel-based density loss term. 
Given the target and source point clouds, we discretize the space into a regular voxel grid and compute occupancy distributions.
The density loss is defined as the mean squared error between the normalized occupancy grids of the target and source point clouds:
\vspace{-0.1cm}
\begin{equation}
    \mathcal{L}_d(\mathcal{P}_\text{trg}, \mathcal{P}_\text{src}) = \frac{1}{V} \sum_{v=1}^V \big( o_v^{(\text{trg})} - o_v^{(\text{src})} \big)^2,
\end{equation}
where $V$ is the total number of voxels, and $o_v^{(\text{trg})}$ and $o_v^{(\text{src})}$ denote the normalized occupancy values in voxel $v$ for target and source respectively.
In sum, minimizing $\mathcal{L}_d$ encourages predicted points to match the global spatial distribution of the ground-truth.

\vspace{0.33em}
\noindent\textbf{Confidence Mask ($\mathcal{L}_m$):}
We supervise the predicted confidence mask $\hat{\mu}$ using binary cross-entropy loss against the corresponding ground-truth valid mask $\mu$, defined as:
\begin{equation}
    \mathcal{L}_m(\mu, \hat{\mu}) = -\frac{1}{N} \sum_{j=1}^{N} \big[ \mu_j \log(\hat{\mu}_j) + (1 - \mu_j) \log(1 - \hat{\mu}_j) \big],
\end{equation}
where $N$ denotes the number of points and $\mu_j \in \{0,1\}$ is the ground-truth validity for point $j$.
Confidence prediction is essential to account for the varying number of points across patches, allowing the network to reconstruct realistic point densities despite the fixed-size padding used for batching.

\section{Experimental Evaluation}
We evaluate our method's capability to efficiently and accurately compress a point cloud scan. We compare against the commonly used codecs: Draco~\cite{draco} and the octree-based compression algorithm from Mekuria et al.~\cite{mekuria2017mpeg}, which we refer to as ``MPEG''. 
Additionally we include the current state-of-the-art approaches RENO~\cite{you2025reno} and OctAttention~\cite{octattention}, as well as RecNet~\cite{stathoulopoulos2024recnet}, that in contrast to ours, encodes each scan into a single latent vector through a range-image embedding. We note that the last three methods, unlike the proposed approach, or Draco/MPEG, do not encode the semantic classes.
\vspace{-0.25cm}

\subsection{Experimental Setup} \label{subsec:setup}
We benchmark our method on the SemanticKITTI~\cite{semkitti} dataset. In order to create the semantic scene graph we use the ground truth semantic labels and in addition we crop each point cloud to 50 meters for all methods, following common practices in the field~\cite{vizzo2023ral}. Similar to SegMap~\cite{dube2020segmap} we use sequences 05 and 06 for training and the rest (00 to 04 and 07 to 10) for the evaluation and comparison to the baselines.

\vspace{0.33em}
\noindent\textbf{Evaluation Metrics:}
The quality of a compression algorithm typically involves a trade-off between compression ratio and reconstruction error. To quantify compression, we use the average bits-per-point (bpp) required to store the encoded point cloud. For our method, this includes all latent vectors as well as the semantic scene graph. 
Following the evaluation protocol in~\cite{wiesmann2021deep}, we assess reconstruction error using three metrics. The first is the symmetric point distance $D_\text{CD}$, defined earlier in Eq.~(\ref{eq:chamfer}) and Eq.~(\ref{eq:distance}), which measures the average nearest-neighbor distance between the reconstructed and ground truth point clouds.
However, for many robotic tasks that rely on point-to-plane ICP registration, reconstructing the exact same points is less critical than ensuring that points lie on the correct surfaces. To address this, we also report the symmetric plane distance $D_\perp$, which is computed similarly to $D_\text{CD}$ but replaces the Euclidean distance $d(\boldsymbol{x}_i,\boldsymbol{x}_j)$ with the point-to-plane distance ${d}_n(\boldsymbol{x}_i,\boldsymbol{x}_j) = |\boldsymbol{n}^\mathsf{T}(\boldsymbol{x}_i - \boldsymbol{x}_j)|$, where $\boldsymbol{n} \in \mathbb{R}^3$ is the ground truth surface normal at the target point, estimated using a $50\,\text{cm}$ neighborhood.
Finally, we include the mean intersection-over-union \textit{(}$IoU$\textit{)} between occupancy grids of the original and reconstructed point clouds, defined as: $IoU = |G_\text{src} \cap G_\text{trg}| \, / \,|G_\text{src} \cup G_\text{trg}|$.
As in~\cite{wiesmann2021deep}, the occupancy grids are computed with a voxel resolution of $20 \times 20 \times 10\,\text{cm}^3$.

\vspace{0.33em}
\noindent\textbf{Implementation Details:}
Our method is implemented in PyTorch. We use distributed training on a set of A100 GPUs and train the model for 150 epochs using the Adam optimizer with a learning rate of $5 \times 10^{-4}$ and a weight decay of $1 \times 10^{-6}$. Each layer is constrained to a max number of points per patch, specifically $\{N_1:720,\,N_2:1720,\,N_3:320,\,N_4:1536\}$.
To vary the compression rate, we evaluate multiple latent vector sizes for each layer. Notably, all other network parameters are kept fixed; only the output dimensionality of each layer is varied across $D_z \in \{8,\,16,\,32,\,64,\,128\}$. The grid size $G$ is set to $2\times2$ for an upsampling factor of 4.
\vspace{-0.25cm}

\subsection{Compression Results and Qualitative Analysis}
\begin{figure}[!h]
    \centering
    \includegraphics[width=1.\linewidth]{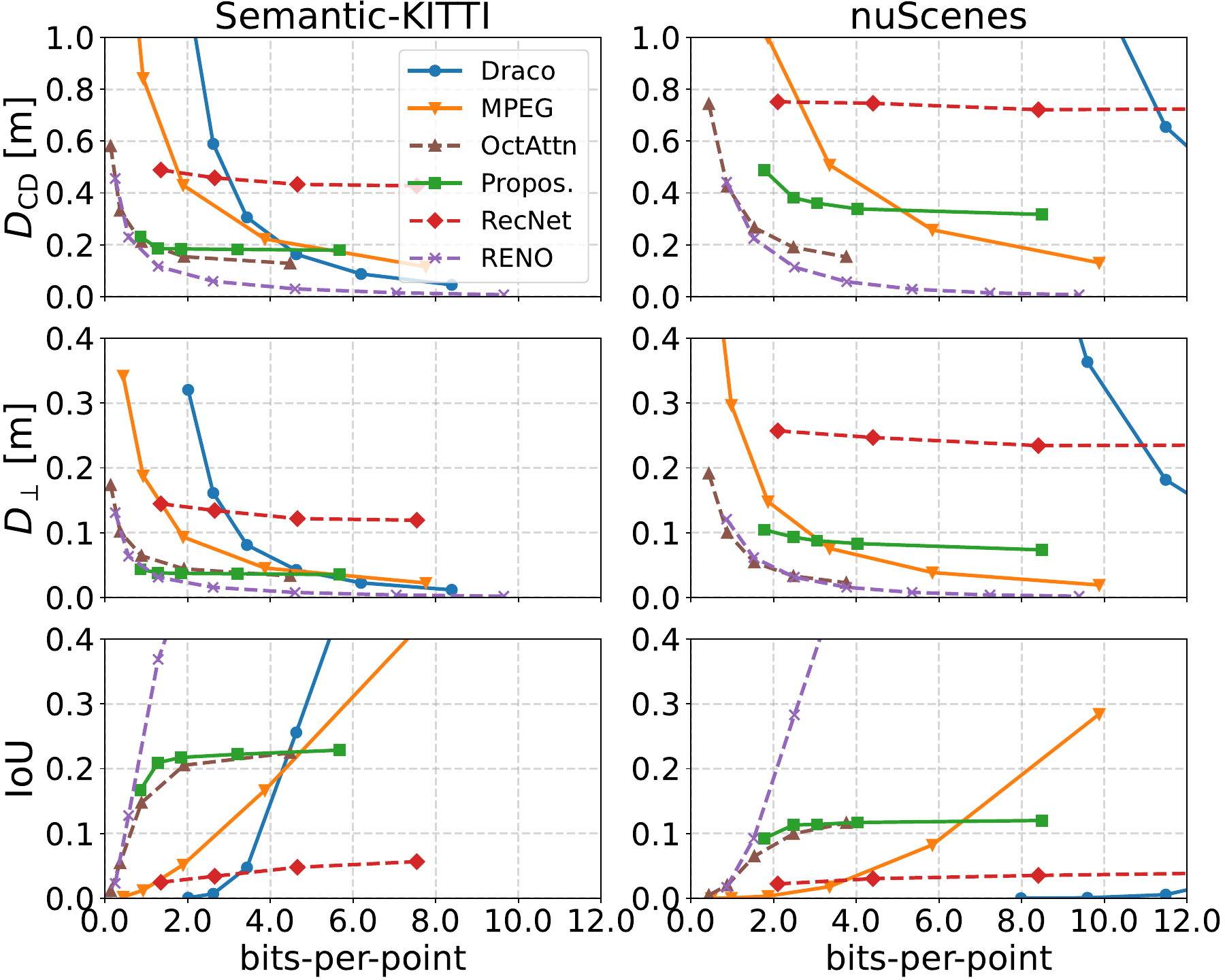}
    \setlength{\abovecaptionskip}{-12pt}
    \caption{\textbf{Compression results per codec.} The left column presents results on SemanticKITTI (in-dataset evaluation), while the right column shows cross-dataset generalization on nuScenes, where models were trained on SemanticKITTI and applied directly to nuScenes without fine-tuning. Dashed lines indicate methods that do not encode semantic labels.}
    \label{fig:compression-results}
    \vspace{-0.5cm}
\end{figure}

We first evaluate the compression performance of our proposed framework in comparison to the established baselines. Our encoding consists of a semantic scene graph whose nodes include both geometric attributes and latent vectors, forming the complete representation required for transmission and decompression. Compression results on SemanticKITTI are presented in the left column of Fig.~\ref{fig:compression-results}, where we vary the latent dimensionality across layers. 
Our method consistently outperforms the classical baselines at lower bits-per-point across all three metrics: $D_\text{CD}$, $D_\perp$, and $IoU$. In particular, in the low bitrate regime ($<4$\,bpp) that is most relevant for bandwidth-limited robotic applications, our approach yields lower $D_\text{CD}$ and $D_\perp$ and higher $IoU$ compared to Draco, MPEG, and RecNet. While MPEG and Draco degrade significantly at low bitrates, and RecNet remains almost flat due to its single-latent design, our method preserves geometric structure and semantic consistency. The slight plateau in performance at higher bitrates is attributed to the fixed network capacity, since only latent size is varied while other architectural parameters remain constant (e.g., number of transformer blocks).

Compared to recent learned codecs, RENO generally achieves the best performance across most operating points, while our method remains consistently competitive with OctAttention. 
In terms of efficiency, OctAttention requires roughly 250\,s per scan, while our method runs at 0.17\,s for encoding and 0.08\,s for decoding on a GeForce RTX~4090 GPU (using 1-3\,GB of VRAM). The generation of the semantic scene graph currently takes about 0.25\,s and remains the main bottleneck in the pipeline. RENO operates in real time (10 FPS). Overall, our approach is competitive with recent learned methods, stronger than the classical codecs and RecNet, and uniquely supports semantic label compression, demonstrating the effectiveness of semantic scene graphs for compression.

\begin{figure*}[!b]
    \centering
    \includegraphics[width=.99\linewidth]{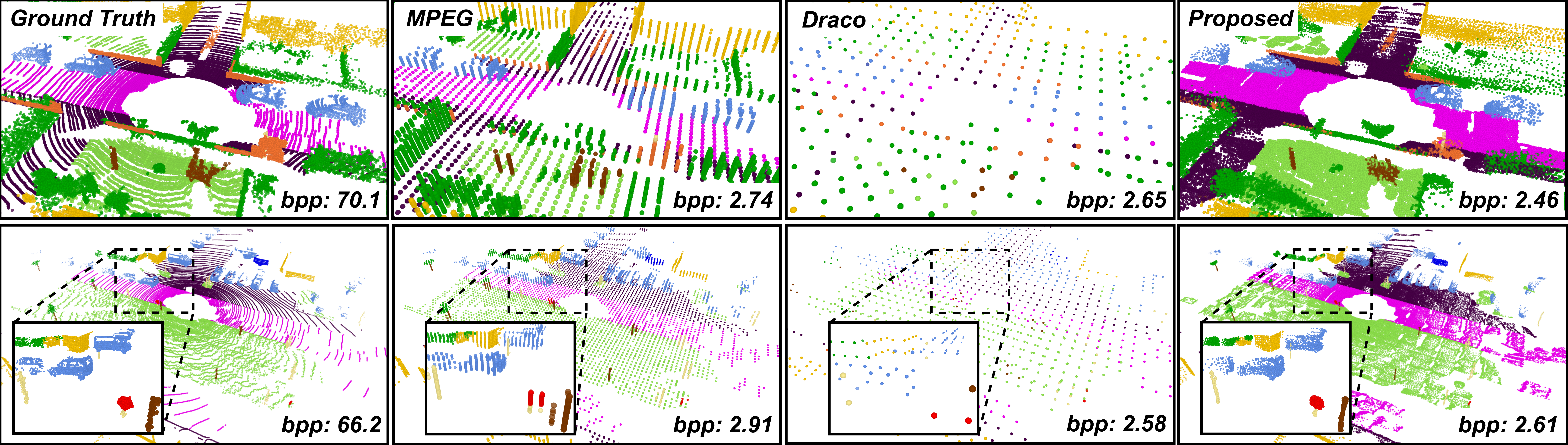}
    \caption{\textbf{Qualitative comparisons.} Qualitative results from the codecs that support semantic label encoding, for two scans of the SemanticKITTI (top: Sequence 00, bottom: Sequence 06), comparing the proposed method with baseline compression algorithms. At low bits-per-point, corresponding to a compression rate of 98\%, our method achieves significantly better reconstruction quality in terms of geometric fidelity and scene completeness. For visual clarity, the ground has been removed in the zoomed-in views, which highlight fine-grained structures such as cars, infrastructure, tree trunks, poles, and signage.}
    \label{fig:reconstruction-examples} 
\end{figure*}
Qualitative reconstructions, shown in Fig.~\ref{fig:reconstruction-examples}, further highlight this trend. At low bits-per-point, baseline methods produce sparse outputs that fail to preserve fine structural details, particularly Draco, which struggles to retain any meaningful geometry due to aggressive quantization. MPEG exhibits better preservation but still suffers from blocky artifacts and loss of surface continuity. In contrast, our method reconstructs scenes with high fidelity, preserving both global layout and local geometry, even at extreme compression levels around 1.8 bpp (compared to 60-70 bpp for the original scans). The results shown correspond to latent dimensions $\{D_{z_1}\!:16,\, D_{z_2}\!:32,\, D_{z_3}\!:16,\, D_{z_4}\!:32\}$, demonstrating the effectiveness of distributed patch-wise encoding.
\vspace{-0.3cm}

\subsection{Generalization Capability}

A common challenge with learning-based methods is poor generalization performance, often displaying significant degradation when applied to environments that differ from the training distribution, primarily due to overfitting. We argue that our method is able to generalize by learning local geometric cues within each semantic layer, rather than learning the whole scene.
To validate this, we deploy the models trained on SemanticKITTI directly to nuScenes without fine-tuning.
This dataset differs substantially from the SemanticKITTI in both sensor configuration (e.g., 32-beam LiDAR with narrower vertical field-of-view and different sensor placement) and geographical context (captured in a different continent, resulting in distinct scene layouts and object appearances).
As shown in the right column of Fig.~\ref{fig:compression-results}, the proposed method maintains better performance compared to baseline approaches at lower bits-per-point. While some degradation in reconstruction accuracy is expected, primarily due to the reduced density of the LiDAR input, our method remains consistently robust across different compression rates. Notably, it also outperforms RecNet, which encodes the entire scene and struggles to generalize to the significantly different distribution of nuScenes. RENO and OctAttention also show strong generalization, with RENO giving the best results overall, while OctAttention remains competitive. In contrast, the baseline methods suffer from drops in both compression efficiency and reconstruction quality, as their dependence on dense local neighborhoods becomes a critical weakness in the sparse and structurally distinct setting of nuScenes.
\vspace{-0.25cm}

\subsection{Ablation Studies}

To validate our design choices, we conducted an ablation study on two key components of our patch-based autoencoder: the semantic conditioning via FiLM and the positional encoding. Table \ref{tab:ablation} summarizes the performance of each variant across layers 2 and 3 using the metrics introduced in~\ref{subsec:setup}.
Disabling both FiLM and the positional encoding led to the most significant degradation in performance, with Chamfer Distance increasing by approximately 21\% and 11\% in layers 2 and 3, and $IoU$ dropping by 19\% and 25\% respectively. This confirms that the two modules provide complementary benefits in enhancing reconstruction fidelity.
Ablating each component independently reveals distinct patterns. Removing FiLM caused a noticeable drop in IoU, particularly in layer 3, highlighting its role in enabling semantic-aware reconstructions, especially for object and agent classes. In contrast, omitting the positional encoding resulted in a higher geometric error across both layers, likely due to the loss of localized spatial priors within the encoder.
These results support our design rationale that FiLM guides the encoder to modulate features based on class semantics, while the positional encoding improves spatial coherence and geometric precision. Their joint use improves both the structural and semantic quality of the reconstructed patches, and is thus retained in our final model.

We further studied the effect of using predicted instead of ground-truth labels for the proposed pipeline, simulating a realistic deployment. Specifically, we tested RangeNet++~\cite{rangenet}, KP-FCNN~\cite{thomas2019KPConv}, and RandLA-Net~\cite{hu2020randla} as semantic segmentation front-ends. The results, reported in Table~\ref{tab:ablation_labels}, show that segmentation accuracy ($mIoU$) is not directly proportional to reconstruction quality. Although RangeNet++ achieves the lowest $mIoU$, it yields the best reconstruction among the three networks, since many of its characteristic \textit{``shadow''} artifacts are filtered out during clustering. In contrast, RandLA-Net, despite comparable $mIoU$, produces larger reconstruction errors due to cluster-level misclassifications, while KP-FCNN performs in between with more localized errors. These findings show that the type and distribution of segmentation errors affect compression performance, offering practical insights for deploying segmentation-compression pipelines in real systems.

\vspace{-0.3cm}
\begin{table}[!b]
\centering
\setlength{\abovecaptionskip}{-0.5pt}
\caption{Ablation study on FiLM and the positional encoding (PosEnc) across Layers 2 \& 3. The symbol (\nmark) indicates the component is used, while the symbol (\xmark) indicates it is ablated.}\label{tab:ablation}
\begin{adjustbox}{width=\columnwidth}
\begin{tabular}{c|c||c|c|c||c|c|c}
\toprule
 \multicolumn{2}{c||}{\textbf{Variant}}  & \multicolumn{3}{c||}{\textbf{Layer 2}} & \multicolumn{3}{c}{\textbf{Layer 3}} \\
\midrule
 \textbf{FiLM} & \textbf{PosEnc} & $\boldsymbol{D_{\text{CD}}}$ $\downarrow$ & $\boldsymbol{D_{\perp}}$ $\downarrow$ & $\boldsymbol{IoU}$ $\uparrow$ & $\boldsymbol{D_{\text{CD}}}$ $\downarrow$ & $\boldsymbol{D_{\perp}}$ $\downarrow$ & $\boldsymbol{IoU}$ $\uparrow$ \\
\midrule
\xmark & \xmark & $0.145$ & $0.047$ & $0.266$ & $0.196$ & $0.068$ & $0.159$ \\
\nmark & \xmark & $0.139$ & $0.042$ & $0.275$ & $0.190$ & $0.059$ & $0.178$ \\
\xmark & \nmark & $0.135$ & $0.039$ & $0.294$ & $0.180$ & $0.054$ & $0.185$ \\
\nmark & \nmark & $\mathbf{0.120}$ & $\mathbf{0.027}$ & $\mathbf{0.330}$ & $\mathbf{0.177}$ & $\mathbf{0.050}$ & $\mathbf{0.211}$\\
\bottomrule
\end{tabular}
\end{adjustbox}
\vspace{-0.4cm}
\end{table}
\begin{table}[!b]
\centering
\setlength{\abovecaptionskip}{-0.5pt}
\caption{Reconstruction performance using predicted vs. ground-truth labels. Relative differences are reported as $\Delta$.}\label{tab:ablation_labels}
\begin{adjustbox}{width=\columnwidth}
\begin{tabular}{l||c|c|c|c|c|c||c}
\toprule
 \textbf{Labeling} & $\boldsymbol{D_{\text{CD}}}$ & $\boldsymbol{D_{\perp}}$ & $\boldsymbol{IoU}$ & $\boldsymbol{\Delta D_{\text{CD}}}$ & $\boldsymbol{\Delta D_{\perp}}$ & $\boldsymbol{\Delta IoU}$ & \textbf{m}$\boldsymbol{IoU}^*$ \\
\midrule
Ground Truth & $0.165$ & $0.034$ & $0.237$ & $-$ & $-$ & $-$ & $-$ \\
\midrule
RangeNet++ & $\mathbf{0.217}$ & $0.036$ & $\mathbf{0.184}$ & $\mathbf{32\%}$ & $6\%$ & $\mathbf{22\%}$ & $52\%$  \\
RandLA-Net & $0.505$ & $0.038$ & $0.045$ & $206\%$ & $11\%$ & $81\%$ & $53\%$ \\
KP-FCNN & $0.253$ & $\mathbf{0.035}$ & $0.131$ & $53\%$ & $\mathbf{3\%}$ & $45\%$ & $\mathbf{58\%}$ \\
\bottomrule
\multicolumn{8}{c}{\vspace{-0.25cm}} \\ 
\multicolumn{8}{c}{$^{*}$The labeling accuracy is presented as the percent mean intersection-over-union over all classes.}
\end{tabular}
\end{adjustbox}
\end{table}

\subsection{Downstream Robotic Tasks}

To validate our compression framework beyond standard metrics, we assess its effectiveness in two representative downstream robotic tasks, as described below.

\vspace{0.33em}
\noindent \textbf{Pose Graph Optimization (PGO):} In a multi-agent setup, we simulate point cloud transmission between agents operating on SemanticKITTI sequences 00 and 07, followed by PGO. Each agent estimates an initial trajectory using \texttt{KISS-ICP}~\cite{vizzo2023ral}
Loop closure candidates between decompressed and raw scans undergo geometric verification via \texttt{GICP}~\cite{fast_gicp}; those with registration error exceeding a predefined threshold are rejected. 
We evaluate two cases where pose graph edges are obtained by registering raw (00) with decompressed (07) scans and vice versa. Results are averaged over both configurations using the \texttt{GTSAM}~\cite{gtsam} optimization.
Performance is measured by the percent improvement in Absolute Trajectory Error (ATE) relative to ground truth. We also quantify the bandwidth required for transmitting raw vs. compressed scans at 10 Hz. As shown in Table~\ref{tab:compression_impact}, MPEG and Draco suffer from high loop rejection and degraded ATE due to poor geometric fidelity, while our method closely follows raw-data trajectory accuracy with roughly 98\% bandwidth reduction.
\begin{table}[!t]
\centering
\setlength{\abovecaptionskip}{-0.5pt}
\caption{Performance of the baseline   and the Proposed (Prop.) Compression Pipeline in Downstream Robotic Tasks}
\label{tab:compression_impact}
\resizebox{\linewidth}{!}{%
\begin{tabular}{l||l||c|c|c|c} 
\toprule
Robotic Tasks & \textbf{Metrics} & \textbf{Raw} & \textbf{MPEG} & \textbf{Draco} & \textbf{Prop.} \\ 
\midrule
\multirow{3}{*}{\shortstack[l]{{Pose Graph Opt.}\\ with \textbf{GTSAM}}} 
& $\uparrow$ Tran. ATE [\%] $^{*}$ & $\mathbf{49.78}$ & $20.23$ & $0.21$ & \underline{$37.76$}\\
& $\uparrow$ Rot. $\,$ATE [\%] $^{*}$ & $\mathbf{23.13}$ & $4.43$ & $0.03$ & \underline{$10.04$}\\
& $\downarrow$ Rej. $\,\,$Loops$^{\dagger}$ & $\mathbf{13}$ & $110$ & $123$ & \underline{$14$}\\
\midrule
\multirow{2}{*}{\shortstack[l]{{Point cloud trans-}\\mission at 10Hz}} 
& $\downarrow$ Bandw. [Mb/s] & $168$ & $3.73$ & $6.04$ & $\mathbf{3.51}$\\
& $\uparrow$ Compres. Rate & N/A & $97.8\%$ & $96.4\%$ & $\mathbf{97.9\%}$ \\
\midrule
\multirow{2}{*}{\shortstack[l]{{Map merging w.}\\\textbf{KISS-Matcher}}} 
& $\downarrow$ Tran. RTE [m] & $\mathbf{0.156}$ & \textit{Failed} & \textit{Failed} & \underline{$0.649$}\\
& $\downarrow$ Rot. $\,$RTE [$\,^\circ$] & $\mathbf{0.862}$ & \textit{Failed} & \textit{Failed} & \underline{$1.310$}\\
\bottomrule
\multicolumn{6}{c}{\vspace{-0.25cm}} \\ 
\multicolumn{6}{c}{$^{*}$The $\%$ improvement in ATE. $^{\dagger}$Rejected Loops after geometric verification.}
\end{tabular}
}
\vspace{-0.25cm}
\end{table}

\begin{figure}[!t]
    \centering
    \setlength{\abovecaptionskip}{-1pt}
    \includegraphics[width=1.0\linewidth]{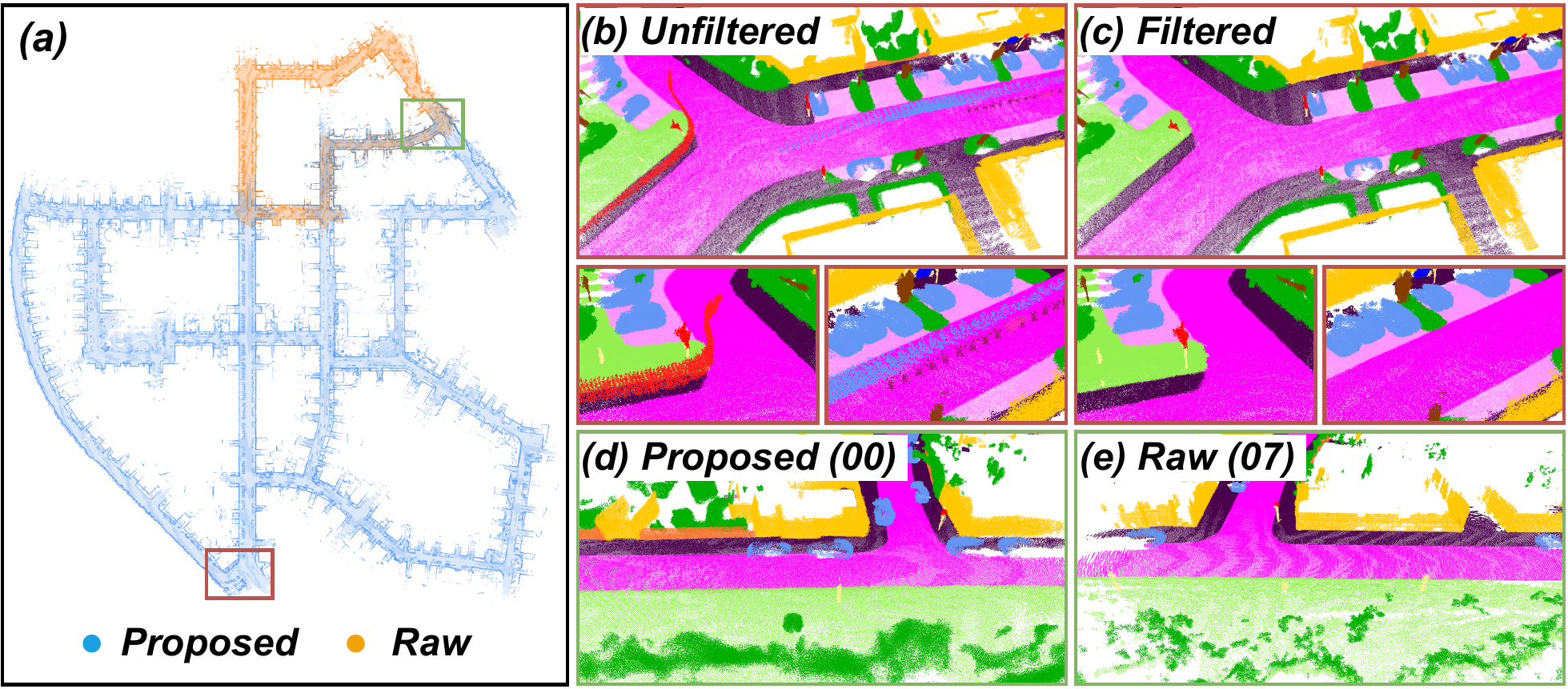}
    \caption{\textbf{Merged maps and zoomed views.} (a) Result of KISS-Matcher~\cite{lim2025icra-KISSMatcher} on KITTI~\cite{semkitti} with the decompressed scans from sequence 00 (using our method) and the raw scans from sequence 07. (b)–(c) Example of the decompressed frames before and after filtering, where dynamic objects (pedestrian, car, motorcyclist) are succesfully removed. (d)–(e) Side-by-side comparison of the overlapping regions: decompressed vs. raw, highlighting preserved structure.} \label{fig:zoomed_results}
\end{figure}
\vspace{-0.2cm}

\vspace{0.33em}
\noindent\textbf{Multi-Robot Map merging:} We further evaluate the utility of our compressed representations in multi-robot map merging. Full-session maps are built from SemanticKITTI sequences using: raw-to-raw (00/07), raw 00 with decompressed 07, and the reverse. Map alignment is performed via \texttt{KISS-Matcher}~\cite{lim2025icra-KISSMatcher}, with mean Relative Transformation Error (RTE) measured against ground truth (see Fig.~\ref{fig:zoomed_results}). Using our scene graph, we apply semantic pruning to remove dynamic obstacles (e.g., pedestrians or vehicles linked to road terrain) from its layered structure, improving clarity as illustrated in Fig.~\ref{fig:zoomed_results}. MPEG and Draco fail to produce reliable correspondences, making them unsuitable for this task, as shown in Table~\ref{tab:compression_impact}. In contrast, our method closely follows the alignment quality of raw data while reducing bandwidth by approximately 98\%.
These results demonstrate that our semantic scene graph-based compression preserves essential geometric and semantic fidelity, while also enabling a practical communication strategy: agents can initially exchange compressed representations, sufficient for tasks such as multi-agent pose graph optimization and map merging, and later, if needed, update with full-resolution scans when higher bandwidth becomes available.
\vspace{-0.1cm}

\section{Discussion and Conclusions}

While our framework achieves strong compression performance, several limitations remain. 
First, generating the semantic scene graph introduces computational overhead that may hinder real-time deployment, and future work will investigate approximate or accelerated graph construction to reduce this bottleneck.
Second, the system depends on a fixed vocabulary from a pre-trained segmentation model, limiting flexibility; integrating open-vocabulary segmentation and dynamic graph generation could enable more adaptable representations. 
For deployment, feasibility on embedded or edge platforms could be improved through model compression techniques (e.g., pruning, quantization, distillation), parallel or task-specific execution of the layer-specific models, and hardware acceleration on embedded GPUs. These directions would reduce latency and memory requirements and make the approach more practical in real-world systems.
In summary, we presented a semantic-aware compression framework that encodes 3D point clouds via patch-wise latent vectors structured by a scene graph. 
Our method achieves state-of-the-art compression under extreme bit-rate constraints while preserving both geometric and semantic structure, and task-driven evaluations show that the representation is effective for downstream robotic applications.

\bibliographystyle{./bibtex/bib/IEEEtran}
\bibliography{./bibtex/bib/IEEEfull,root}

@inproceedings{GICP,
  title={{Generalized-ICP}},
  author={Aleksandr V. Segal and Dirk H{\"a}hnel and Sebastian Thrun},
  booktitle={Robotics: Science and Systems},
  year={2009}
}

@article{dube2020segmap,
    author = {Renaud Dubé and Andrei Cramariuc and Daniel Dugas and Hannes Sommer and Marcin Dymczyk and Juan Nieto and Roland Siegwart and Cesar Cadena},
    title ={{SegMap: Segment-based mapping and localization using data-driven descriptors}},
    journal = {The International Journal of Robotics Research},
    volume = {39},
    number = {2-3},
    pages = {339-355},
    year = {2020},
    doi = {10.1177/0278364919863090},
}

@inproceedings{attentionTrans,
author = {Vaswani, Ashish and Shazeer, Noam and Parmar, Niki and Uszkoreit, Jakob and Jones, Llion and Gomez, Aidan N. and Kaiser, \L{}ukasz and Polosukhin, Illia},
title = {Attention is All You Need},
year = {2017},
isbn = {9781510860964},
publisher = {Curran Associates Inc.},
booktitle = {Proceedings of the 31st International Conference on Neural Information Processing Systems (NIPS)},
pages = {6000–6010},
numpages = {11},
}

@article{damigos2023com_aware,
    author={Damigos, Gerasimos
    and Stathoulopoulos, Nikolaos
    and Koval, Anton
    and Lindgren, Tore
    and Nikolakopoulos, George},
    title={{Communication-Aware Control of Large Data Transmissions via Centralized Cognition and 5G Networks for Multi-Robot Map merging}},
    journal={Journal of Intelligent {\&} Robotic Systems},
    year={2024},
    month={Jan},
    day={10},
    volume={110},
    number={1},
    pages={22},
    issn={1573-0409},
    doi={10.1007/s10846-023-02045-4}
  }

@inproceedings{stathoulopoulos2024recnet,
    title={{RecNet: An Invertible Point Cloud Encoding through Range Image Embeddings for Multi-Robot Map Sharing and Reconstruction}}, 
    author={Nikolaos Stathoulopoulos and Mario A. V. Saucedo and Anton Koval and George Nikolakopoulos},
    year={2024},
    eprint={2402.02192},
    archivePrefix={arXiv},
    primaryClass={cs.RO},
    booktitle={IEEE International Conference on Robotics and Automation (ICRA)},
}

@inproceedings{Theis2022riddle,
  author = {Theis, Lucas and Shi, Wenzhe},
  booktitle = {IEEE / CVF Computer Vision and Pattern Recognition Conference 2022},
  pages = {1--19},
  title = {{RIDDLE: Lidar Data Compression with Range Image Deep Delta Encoding}},
  year = {2022}
}

@article{Feng2020,
  author = {Feng, Yu and Liu, Shaoshan and Zhu, Yuhao},
  doi = {10.1109/IROS45743.2020.9341071},
  isbn = {9781728162126},
  issn = {21530866},
  journal = {IEEE International Conference on Intelligent Robots and Systems},
  pages = {10766--10773},
  title = {{Real-time spatio-temporal LiDAR point cloud compression}},
  year = {2020}
}

@ARTICLE{stathoulopoulos2024frame,
  author={Stathoulopoulos, Nikolaos and Lindqvist, Björn and Koval, Anton and Agha-Mohammadi, Ali-Akbar and Nikolakopoulos, George},
  journal={IEEE Transactions on Field Robotics},
  title={{FRAME: A Modular Framework for Autonomous Map Merging: Advancements in the Field}},
  year={2024},
  volume={1},
  number={},
  pages={1-26},
  doi={10.1109/TFR.2024.3419439}
}

@INPROCEEDINGS{fast_gicp,
  author={Koide, Kenji and Yokozuka, Masashi and Oishi, Shuji and Banno, Atsuhiko},
  booktitle={2021 IEEE International Conference on Robotics and Automation (ICRA)}, 
  title={{Voxelized GICP for Fast and Accurate 3D Point Cloud Registration}}, 
  year={2021},
  volume={},
  number={},
  pages={11054-11059},
  doi={10.1109/ICRA48506.2021.9560835}}

@ARTICLE{Sun2019clustering,
  author={Sun, Xuebin and Ma, Han and Sun, Yuxiang and Liu, Ming},
  journal={IEEE Robotics and Automation Letters}, 
  title={{A Novel Point Cloud Compression Algorithm Based on Clustering}}, 
  year={2019},
  volume={4},
  number={2},
  pages={2132-2139},
  doi={10.1109/LRA.2019.2900747}}

@ARTICLE{Zhao2022semantic,
  author={Zhao, Lili and Ma, Kai-Kuang and Liu, Zhili and Yin, Qian and Chen, Jianwen},
  journal={IEEE Transactions on Circuits and Systems for Video Technology}, 
  title={{Real-Time Scene-Aware LiDAR Point Cloud Compression Using Semantic Prior Representation}}, 
  year={2022},
  volume={32},
  number={8},
  pages={5623-5637},
  doi={10.1109/TCSVT.2022.3145513}}

@INPROCEEDINGS{Que2021VoxelContext,
  author={Que, Zizheng and Lu, Guo and Xu, Dong},
  booktitle={2021 IEEE/CVF Conference on Computer Vision and Pattern Recognition (CVPR)}, 
  title={{VoxelContext-Net: An Octree based Framework for Point Cloud Compression}}, 
  year={2021},
  volume={},
  number={},
  pages={6038-6047},
  doi={10.1109/CVPR46437.2021.00598}}

@INPROCEEDINGS{Heo2022FLiCR,
  author={Heo, Jin and Phillips, Christopher and Gavrilovska, Ada},
  booktitle={2022 IEEE/ACM 7th Symposium on Edge Computing (SEC)}, 
  title={{FLiCR: A Fast and Lightweight LiDAR Point Cloud Compression Based on Lossy RI}}, 
  year={2022},
  volume={},
  number={},
  pages={54-67},
  doi={10.1109/SEC54971.2022.00012}}

@ARTICLE{Wang2023tensor,
  author={Wang, Jianqiang and Ding, Dandan and Li, Zhu and Feng, Xiaoxing and Cao, Chuntong and Ma, Zhan},
  journal={IEEE Transactions on Pattern Analysis and Machine Intelligence}, 
  title={{Sparse Tensor-Based Multiscale Representation for Point Cloud Geometry Compression}}, 
  year={2023},
  volume={45},
  number={7},
  pages={9055-9071},
  doi={10.1109/TPAMI.2022.3225816}}

@article{wiesmann2021deep,
  title={{Deep Compression for Dense Point Cloud Maps}},
  author={Wiesmann, Louis and Milioto, Andres and Chen, Xieyuanli and Stachniss, Cyrill and Behley, Jens},
  journal={IEEE Robotics and Automation Letters},
  volume={6},
  number={2},
  pages={2060--2067},
  year={2021},
  publisher={IEEE}
}

@inproceedings{you2025reno,
  title={{RENO: Real-time Neural Compression for 3D LiDAR Point Clouds}},
  author={You, Kang and Chen, Tong and Ding, Dandan and Asif, M Salman and Ma, Zhan},
  booktitle={Proceedings of the Computer Vision and Pattern Recognition Conference},
  pages={22172--22181},
  year={2025}
}

@ARTICLE{mekuria2017mpeg,
  author={Mekuria, Rufael and Blom, Kees and Cesar, Pablo},
  journal={IEEE Transactions on Circuits and Systems for Video Technology}, 
  title={{Design, Implementation, and Evaluation of a Point Cloud Codec for Tele-Immersive Video}}, 
  year={2017},
  volume={27},
  number={4},
  pages={828-842},
  doi={10.1109/TCSVT.2016.2543039}}

@article{wang2019dynamic,
    author = {Wang, Yue and Sun, Yongbin and Liu, Ziwei and Sarma, Sanjay E. and Bronstein, Michael M. and Solomon, Justin M.},
    title = {{Dynamic Graph CNN for Learning on Point Clouds}},
    year = {2019},
    issue_date = {October 2019},
    publisher = {Association for Computing Machinery},
    address = {New York, NY, USA},
    volume = {38},
    number = {5},
    issn = {0730-0301},
    doi = {10.1145/3326362},
    journal = {ACM Trans. Graph.},
    month = oct,
    articleno = {146},
    numpages = {12}
}

@inproceedings{hu2020randla,
  title={{RandLA-Net: Efficient Semantic Segmentation of Large-scale Point Clouds}},
  author={Hu, Qingyong and Yang, Bo and Xie, Linhai and Rosa, Stefano and Guo, Yulan and Wang, Zhihua and Trigoni, Niki and Markham, Andrew},
  booktitle={Proceedings of the IEEE/CVF Conference on Computer Vision and Pattern Recognition},
  pages={11108--11117},
  year={2020}
}

@inproceedings{armeni20193d,
  title={{3D Scene Graph: A Structure for Unified Semantics, 3D Space, and Camera}},
  author={Armeni, Iro and He, Zhi-Yang and Gwak, JunYoung and Zamir, Amir R and Fischer, Martin and Malik, Jitendra and Savarese, Silvio},
  booktitle={Proceedings of the IEEE International Conference on Computer Vision},
  pages={5664--5673},
  year={2019}
}

@article{gao2023scenehgn,
    author = {Gao, Lin and Sun, Jia-Mu and Mo, Kaichun and Lai, Yu-Kun and Guibas, Leonidas J. and Yang, Jie},
    title = {{SceneHGN: Hierarchical Graph Networks for 3D Indoor Scene Generation With Fine-Grained Geometry}},
    year = {2023},
    issue_date = {July 2023},
    publisher = {IEEE Computer Society},
    address = {USA},
    volume = {45},
    number = {7},
    issn = {0162-8828},
    doi = {10.1109/TPAMI.2023.3237577},
    journal = {IEEE Trans. Pattern Anal. Mach. Intell.},
    month = jul,
    pages = {8902–8919},
    numpages = {18}
}

@InProceedings{perez2018film,
  title={{FiLM: Visual Reasoning with a General Conditioning Layer}},
  author={Ethan Perez and Florian Strub and Harm de Vries and Vincent Dumoulin and Aaron C. Courville},
  booktitle={AAAI},
  year={2018}
}

@INPROCEEDINGS{foldingnet,
  author={Yang, Yaoqing and Feng, Chen and Shen, Yiru and Tian, Dong},
  booktitle={2018 IEEE/CVF Conference on Computer Vision and Pattern Recognition}, 
  title={{FoldingNet: Point Cloud Auto-Encoder via Deep Grid Deformation}}, 
  year={2018},
  volume={},
  number={},
  pages={206-215},
  doi={10.1109/CVPR.2018.00029}}

@inproceedings{draco,
    author = {Google},
    title = {{Draco 3D Data Compression}},
    year = {2017},
    url = {https://github.com/google/draco},
}

@INPROCEEDINGS{damigos2024-5G,
  author={Damigos, Gerasimos and Saradagi, Akshit and Sandberg, Sara and Nikolakopoulos, George},
  booktitle={2024 IEEE International Conference on Robotics and Automation (ICRA)}, 
  title={{Environmental Awareness Dynamic 5G QoS for Retaining Real Time Constraints in Robotic Applications}}, 
  year={2024},
  volume={},
  number={},
  pages={12069-12075},
  doi={10.1109/ICRA57147.2024.10610698}
}

@inproceedings{lim2025icra-KISSMatcher,
  author={Lim, Hyungtae and Kim, Daebeom and Shin, Gunhee and Shi, Jingnan and Vizzo, Ignacio and Myung, Hyun and Park, Jaesik and Carlone, Luca},
  booktitle={2025 IEEE International Conference on Robotics and Automation (ICRA)}, 
  title={{KISS-Matcher: Fast and Robust Point Cloud Registration Revisited}}, 
  year={2025},
  volume={},
  number={},
  pages={11104-11111},
  doi={10.1109/ICRA55743.2025.11127458}
}

@article{vizzo2023ral,
  author    = {Vizzo, Ignacio and Guadagnino, Tiziano and Mersch, Benedikt and Wiesmann, Louis and Behley, Jens and Stachniss, Cyrill},
  title     = {{KISS-ICP: In Defense of Point-to-Point ICP -- Simple, Accurate, and Robust Registration If Done the Right Way}},
  journal   = {IEEE Robotics and Automation Letters (RA-L)},
  pages     = {1029--1036},
  doi       = {10.1109/LRA.2023.3236571},
  volume    = {8},
  number    = {2},
  year      = {2023},
  codeurl   = {https://github.com/PRBonn/kiss-icp},
}

@inproceedings{semkitti,
  author = {J. Behley and M. Garbade and A. Milioto and J. Quenzel and S. Behnke and C. Stachniss and J. Gall},
  title = {{SemanticKITTI: A Dataset for Semantic Scene Understanding of LiDAR Sequences}},
  booktitle = {2019 IEEE/CVF International Conference on Computer Vision (ICCV)},
  year = {2019}
}

@software{gtsam,
  author       = {Frank Dellaert and GTSAM Contributors},
  title        = {borglab/gtsam},
  month        = May,
  year         = 2022,
  publisher    = {Georgia Tech Borg Lab},
  version      = {4.2a8},
  doi          = {10.5281/zenodo.5794541},
  url          = {https://github.com/borglab/gtsam}
}

@ARTICLE{seisa2022edge,
  author={Seisa, Achilleas Santi and Lindqvist, Björn and Satpute, Sumeet Gajanan and Nikolakopoulos, George},
  journal={IEEE Access}, 
  title={{E-CNMPC: Edge-Based Centralized Nonlinear Model Predictive Control for Multiagent Robotic Systems}}, 
  year={2022},
  volume={10},
  number={},
  pages={121590-121601},
  doi={10.1109/ACCESS.2022.3223446}}

@article{octattention, 
  title={{OctAttention: Octree-Based Large-Scale Contexts Model for Point Cloud Compression}},
  volume={36},
  DOI={10.1609/aaai.v36i1.19942},
  number={1},
  journal={Proceedings of the AAAI Conference on Artificial Intelligence},
  author={Fu, Chunyang and Li, Ge and Song, Rui and Gao, Wei and Liu, Shan},
  year={2022},
  month={Jun.},
  pages={625-633}
}

@article{thomas2019KPConv,
    Author = {Thomas, Hugues and Qi, Charles R. and Deschaud, Jean-Emmanuel and Marcotegui, Beatriz and Goulette, Fran{\c{c}}ois and Guibas, Leonidas J.},
    Title = {{KPConv: Flexible and Deformable Convolution for Point Clouds}},
    Journal = {Proceedings of the IEEE International Conference on Computer Vision},
    Year = {2019}
}

@INPROCEEDINGS{rangenet,
  author={Milioto, Andres and Vizzo, Ignacio and Behley, Jens and Stachniss, Cyrill},
  booktitle={2019 IEEE/RSJ International Conference on Intelligent Robots and Systems (IROS)}, 
  title={{RangeNet++: Fast and Accurate LiDAR Semantic Segmentation}}, 
  year={2019},
  volume={},
  number={},
  pages={4213-4220},
  doi={10.1109/IROS40897.2019.8967762}}

\end{document}